\pdfoutput=1

\documentclass[11pt]{article}

\usepackage[usenames,dvipsnames]{xcolor}
\usepackage[preprint]{acl}

\usepackage{graphicx}
\usepackage{times}
\usepackage{latexsym}
\usepackage{amsmath}
\usepackage{cleveref}

\usepackage[T1]{fontenc}

\usepackage[utf8]{inputenc}

\usepackage{microtype}

\usepackage{inconsolata}
\usepackage{nicematrix}

\newcount\Comments  
\newcommand{\kibitz}[2]{\ifnum\Comments=1\textcolor{#1}{#2}\fi}

\newcommand{\nico}[1]{\kibitz{orange}     {\small{[Nico: #1]}}}

\definecolor{tablecolor}{rgb}{0.8,0.8,0.8}

\newcommand{\vparam}{\vtheta}

\newcommand\cut[1]{}

\newcommand{\squishlist}{
   \begin{list}{$\bullet$}
    { \setlength{\itemsep}{0pt}      \setlength{\parsep}{3pt}
      \setlength{\topsep}{3pt}       \setlength{\partopsep}{0pt}
      \setlength{\leftmargin}{1.5em} \setlength{\labelwidth}{1em}
      \setlength{\labelsep}{0.5em} } }

\newcommand{\squishlisttwo}{
   \begin{list}{$\bullet$}
    { \setlength{\itemsep}{0pt}    \setlength{\parsep}{0pt}
      \setlength{\topsep}{0pt}     \setlength{\partopsep}{0pt}
      \setlength{\leftmargin}{2em} \setlength{\labelwidth}{1.5em}
      \setlength{\labelsep}{0.5em} } }

\newcommand{\squishend}{
    \end{list}  }

{}
{}
{}

\newcommand{\myvec}[1]{\mbox{$\mathbf{#1}$}}
\newcommand{\myvecsym}[1]{\mbox{$\boldsymbol{#1}$}}

\newcommand{\vpsi}{\myvecsym{\psi}}

\newcommand{\vtheta}{\mbox{$\myvecsym{\theta}$}}

\newcommand{\vr}{\mbox{$\myvec{r}$}}
\newcommand{\vs}{\mbox{$\myvec{s}$}}
\newcommand{\vt}{\mbox{$\myvec{t}$}}

\newcommand{\calD}{\mbox{${\cal D}$}}

\newcommand{\data}{\calD}

\usepackage{colortbl}
\usepackage{times}
\usepackage{latexsym}
\usepackage{microtype}
\usepackage{bbm}
\usepackage{todonotes}
\usepackage{amsmath}
\usepackage{amsfonts}
\usepackage{adjustbox,lipsum}
\usepackage{tabularx}
\usepackage{booktabs}   
\usepackage{caption}
\usepackage{subcaption}
\usepackage{xspace}
\usepackage{enumitem}
\usepackage{arydshln}
\usepackage[most]{tcolorbox}
\usepackage{quoting}
\usepackage{listings}
\title{Socratic Reasoning Improves Positive Text Rewriting}

\author{
    Anmol Goel$^{a}$,
    Nico Daheim$^{a}$, 
    Christian Montag$^{b}$,
    Iryna Gurevych$^{a}$
 \\ 
  $^{a}$Ubiquitous Knowledge Processing Lab (UKP Lab) \\
Department of Computer Science and Hessian Center for AI (hessian.AI) \\
Technical University of Darmstadt
  \\
  $^{b}$Centre for Cognitive and Brain Sciences, \\ Institute of Collaborative Innovation, \\
  University of Macau, Macau, China
\\
  \url{www.ukp.tu-darmstadt.de}
}

\begin{document}
\maketitle
\begin{abstract}
Reframing a negative into a positive thought is at the crux of several cognitive approaches to mental health and psychotherapy that could be made more accessible by large language model-based solutions.
Such reframing is typically non-trivial and requires multiple rationalization steps to uncover the underlying issue of a negative thought and transform it to be more positive.
However, this rationalization process is currently neglected by both datasets and models which reframe thoughts in one step.
In this work, we address this gap by augmenting open-source datasets for positive text rewriting with synthetically-generated Socratic rationales using a novel framework called \textsc{SocraticReframe}. 
\textsc{SocraticReframe} uses a sequence of question-answer pairs to rationalize the thought rewriting process.
We show that such Socratic rationales significantly improve positive text rewriting for different open-source LLMs according to both automatic and human evaluations guided by criteria from psychotherapy research. We validate our framework and the synthetic rationalizations with expert judgements from domain experts and psychology students in an IRB-approved annotation study. Our findings  highlight the potential of utilizing the synergy between LLM reasoning and established psychotherapy techniques to build assistive solutions for reframing negative thoughts.
\end{abstract}

\section{Introduction}

Negative thoughts can have a profound effect on human judgement and well-being.
Oftentimes negative thoughts overshadow positive thoughts \cite{vaish2008not}, because they can be emotionally deep-rooted and triggering when brought to surface~\cite{beck1979cognitive}.
\textit{Cognitive Reframing} is a highly-validated Cognitive Behavioral Therapy (CBT) intervention technique that aims to address this by identifying and reframing negative thoughts into positive ones \cite{clark2013cognitive} and has proven useful in both clinical therapeutic and self-help settings \cite{williams2001use}. Therefore, reframing holds great potential as a strategy for clinicians to provide help in overcoming negative thoughts.
Beyond this, teaching people to reframe their own thoughts as a self-guided mental health intervention could provide them with a coping strategy that does not rely on clinical help \citep{jorm2006effectiveness}. 
However, coming up with effective reframes can be challenging and training people, both for clinical and self-help settings, to effectively reframe negative thoughts is a time-consuming and laborious process. 
Altogether, this poses a risk of mental health improvements being slowed by a lack of availability of knowledge and clinicians.

\begin{figure}[!t]
    \centering
    \includegraphics[width=\columnwidth]{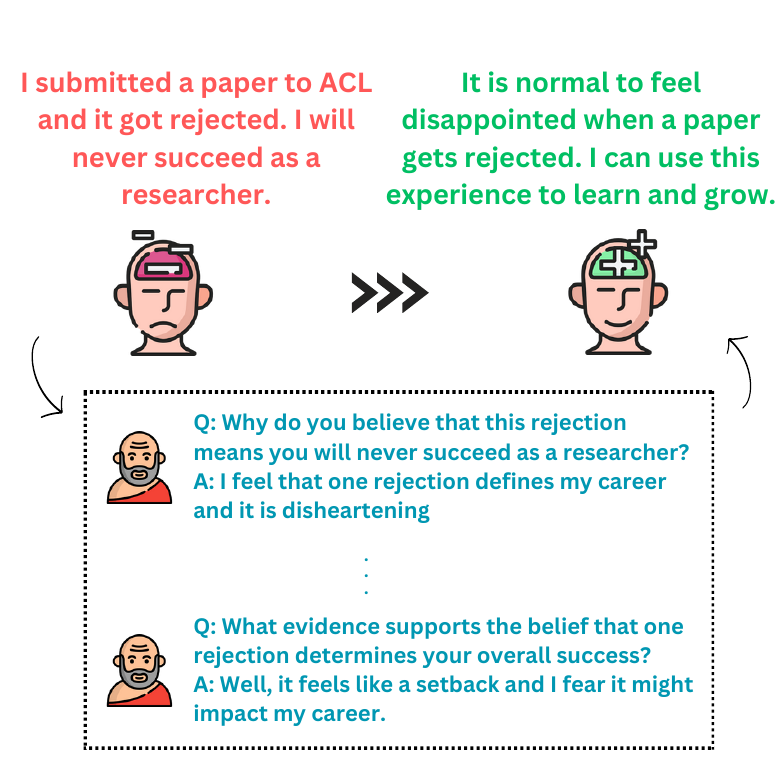}
    \caption{We use Socratic rationales consisting of question-answer pairs to improve positive text rewriting by verbalising the rewriting process. Here, we illustrate this: given a \textcolor{RedOrange}{negative thought}, we generate \textcolor{BlueGreen}{Socratic rationales} grounded in Cognitive Behavioral Therapy to reframe the original thought into a \textcolor{OliveGreen}{positive thought}.}
    \label{fig:enter-label}
\end{figure}

Large Language Models (LLMs) hold great promise to overcome this lack of clinician support and self-help resources by providing assistive solutions and could also be used as a training resource.
Notably, it has been shown that they can effectively guide text generation towards desired attributes like non-toxicity \cite{liu-etal-2021-dexperts,zheng-etal-2023-click}, style \cite{reif-etal-2022-recipe,tikhonov-etal-2019-style}, or persona \cite{song-etal-2020-generate}, and could therefore provide a diverse set of reframes.
This motivates this work, where we aim to improve the reframing performance of LLMs. 

While LLMs hold great promise for mental health applications due to their step-by-step reasoning and rationalization capabilities~\citep[inter alia]{kojima2022large}, they require careful investigation in tandem with domain experts. 
Several factors need to be taken into consideration before deploying LLM-based applications for client or therapist-facing applications. 
LLM-based solutions should not only be consistent with the client's mental state, interpretable and avoid hallucinations, but importantly should also follow CBT guidelines.

Therefore, we aim to tackle this gap in literature by making the reasoning of LLMs explicit and include it as a supervision signal during training. We train LLMs for the positive text rewriting task by first generating a rationalization using the Socratic method\footnote{The terminology for the Socratic method is not consistent within the literature \cite{carey2004socratic}. 
In the context of this work, we use the term Socratic reasoning to refer to question-answer sequences that follow the Socratic method.}, which uses questioning strategies for uncovering a person's beliefs and motivations, and has shown to be a useful tool for CBT-based cognitive reframing. Socratic reasoning typically uses probing questions to elicit alternative perspectives. \Cref{tab:socratic_types} lists different types of Socratic questions that are common in literature and practice. 
We call our method \textsc{SocraticReframe}.

We benchmark \textsc{SocraticReframe} with various state-of-the-art LLMs for the task of positive text rewriting using both automatic and human evaluations with clinical experts.
Our results show that explicitly using Socratic reasoning as a training signal improves the text rewriting performance of LLMs while staying interpretable and faithful to CBT guidelines. We conduct extensive analysis on the synthetically-generated Socratic rationales and find that the socratic rationales are both informative to the model and adhere to clinical standards according to a known Socratic-questioning evaluation scheme.
Here, the rationales are rated as highly helpful and relevant for the reframing process.

We use the generated rationales to augment three existing cognitive reframing datasets from related works. We release all data and code publicly to enable subsequent research in the mental health domain, for example, for clinician training. 

\section{Background \& Related Work} \label{background}
\begin{table*}[!t]
\centering
\small
\begin{tabular}{p{4cm}p{6cm}p{5cm}}
\hline
\textbf{Question Type} & \textbf{Description} & \textbf{Exemplars}\\
\hline
Clarification & Questions to go deeper into a thought & ``Why do you say that?'' \\ 
Probing assumptions & Questions to make someone think about unquestioned beliefs & ``What could we assume instead?'' \\
Probing reasons and evidence & Questions digging into the reasoning behind a thought & ``How did you know that...?'' \\ 
Probing implications & Questions probing the consequences of a thought & ``What is likely to happen if ...?'' \\ 
Probing alternative viewpoints & Questions about other, equally valid viewpoints & ``What is another way to look at it?'' \\
Question about the question & Meta-questions about the question itself & ``Why do you think I asked this?'' \\
\hline
\end{tabular}
\caption{\label{tab:socratic_types}
The six types of Socratic questions with representative exemplars from \cite{paul2019thinker}. We synthetically generate Socratic questions spanning all types to improve cognitive reframing with (\Cref{socraticreframe}).
}
\end{table*}
\subsection{NLP and Mental Health}
Research on NLP for mental health has primarily focused on utilizing linguistic features, as well as neural representations, to identify and analyze mental health conditions such as depression \cite{rinaldi-etal-2020-predicting, yates-etal-2017-depression}, anxiety \cite{juhng-etal-2023-discourse, wei-etal-2021-linguistic}, dyslexia \cite{bjornsdottir-etal-2023-dyslexia, gala-ziegler-2016-reducing}, autism \cite{cho-etal-2022-identifying, goodkind-etal-2018-detecting}, and schizophrenia \cite{mitchell-etal-2015-quantifying,sarioglu-kayi-etal-2017-predictive}, among others. Typically, these studies rely on crowdsourced data or annotated social media posts to address the ethical and privacy concerns associated with medical data \cite{harrigian-etal-2020-models,mossburger-etal-2020-exploring,turcan-mckeown-2019-dreaddit}. Recently, efforts have been made to utilize datasets that are more representative of the interactions between a therapist and client in real world settings \cite{perez-rosas-etal-2017-understanding,shapira-etal-2022-measuring,howes-etal-2014-linguistic,lee-etal-2019-identifying,cao-etal-2019-observing,tanana-etal-2015-recursive,shreevastava-foltz-2021-detecting}. Additionally, recent works have utilized synthetic data to enhance the performance of models in clinical \cite{kazi-kahanda-2019-automatically,hiebel-etal-2023-synthetic,shim-etal-2021-synthetic,lindsay-etal-2022-generating} and mental health \cite{wu-etal-2023-experts} settings. \citet{chen2023empowering} propose Diagnosis-of-Thought prompting focusing only on diagnosing the type of cognitive distortion given a patient's speech. In contrast, we develop a novel training framework for improving LLM performance on cognitive reframing. Our work aims to contribute to the growing body of literature leveraging positive psychology \cite{sheng-etal-2023-learning} and LLMs to improve performance on mental health tasks. 

\subsection{Socratic Questioning}
The Socratic method has found wide use in pedagogy \cite{bautista2014socratic} and psychotherapy \cite{braun2015therapist}, because it can improve understanding and enable alternative perspectives without being explicit or direct. 
Instead, it uses questioning strategies to leave room for exploration which is also helpful for other applications, such as tutoring~\citep{macina-etal-2023-opportunities}.
Recent works have used the step-by-step nature of Socratic questioning to improve NLP methods. \citet{ang-etal-2023-socratic} collect a large-scale dataset from Reddit annotated with a question type to train Socratic question generation models. \citet{wu-etal-2023-hence} develop a benchmark of Socratic-inspired deductive reasoning patterns to train state-of-the-art textual entailment and question answering models. \citet{pagnoni-etal-2023-socratic} propose Socratic pretraining to enable a question-driven approach for summarizing documents. The scaffolding of Socratic questioning also enables LLMs to solve complex problems by decomposing them into smaller sub-problems. \citet{qi-etal-2023-art, shridhar-etal-2022-automatic, shridhar-etal-2023-distilling} use Socratic methods to improve LLM's performance on a variety of reasoning tasks, including math word problem solving and logical reasoning. Our work uses the Socratic method as the core principle to guide a language model while generating a reframed thought.


\section{\textsc{SocraticReframe}} \label{socraticreframe}
\begin{figure*}[t!]
    \centering
    \includegraphics[width=\textwidth]{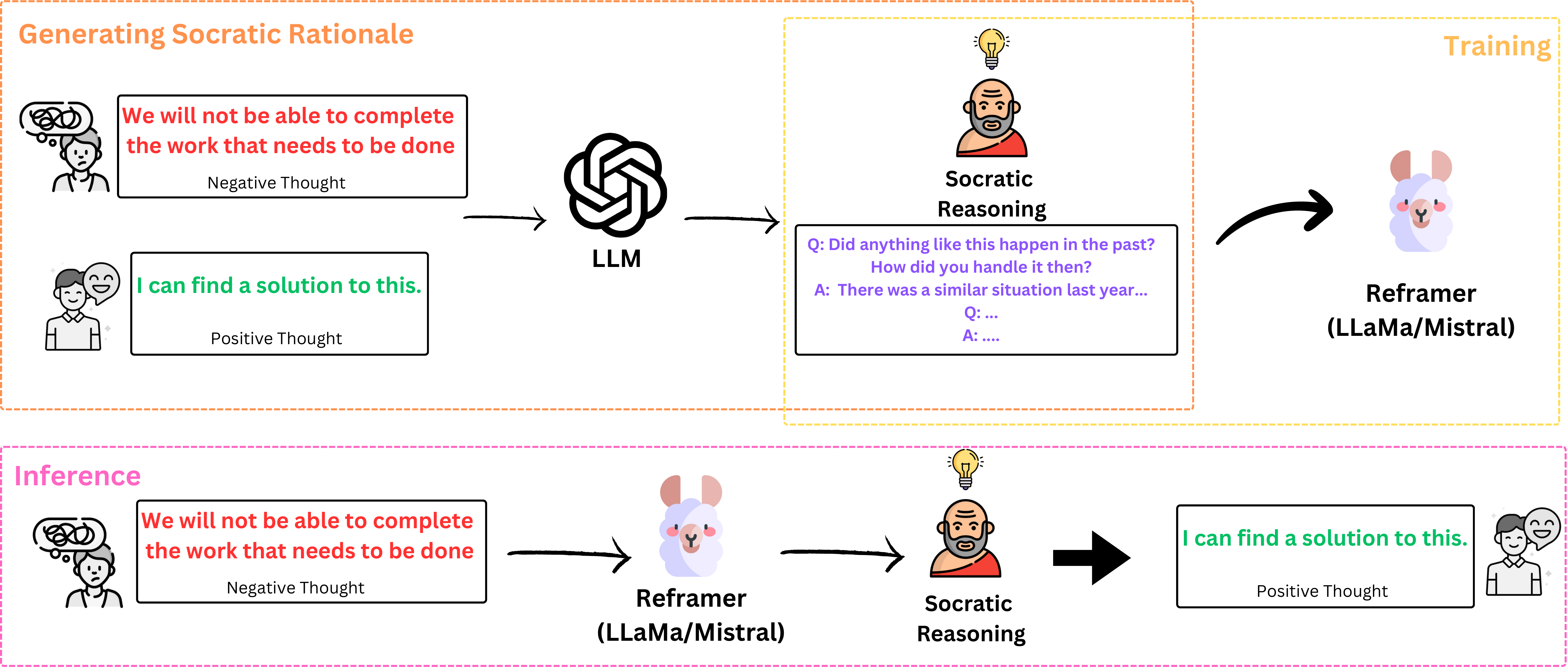}
    \caption{Detailed illustration of our framework \textsc{SocraticReframe}. First, we generate Socratic rationales using GPT-4 with a few-shot prompt. Then, we use the generated Socratic rationale to train models for cognitive reframing. During inference, the model generates the Socratic rationale before reframing the negative thought.}
    \label{fig:full_framework}
\end{figure*}
Cognitive reframing aims to reframe a negative thought into a positive thought.
We approach this with LLMs that generate a positive thought conditioned on the negative thought and metadata.
However, directly training an LLM to map positive to negative thought does not make the process behind the reframing explicit, which is important both for interpretability and model performance.
Our method \textsc{SocraticReframe} overcomes this by training the model to verbalize a Socratic rationale before generating the reframing.

Formally, cognitive reframing transforms a negative thought $\vt \in \mathcal{V}^\ast$ into a positive thought $\vr \in \mathcal{V}^\ast$. 
Both are given as strings constructed from a (model) vocabulary $\mathcal{V}$.
In addition to the negative thought, metadata $\vpsi \in \mathcal{V}^\ast$ is often available which, for example, describes the person experiencing distress and the situation that caused it.
One method for automatic reframing is using an autoregressive LLM with parameters $\vparam$ to model the distribution
\begin{equation}
\begin{split}
p_\text{\vparam}(\vr\mid \vt, \vpsi) &= \prod_{n=1}^{|\vr|}p_\text{\vparam}(r_n \mid \vt, \vpsi, \vr_{<n})    
\end{split}
\end{equation}
paired with a decoding strategy, such as, sampling or greedy decoding.
The model can be both finetuned using paired data $\smash{\data_{\text{train}} = \{(\vt_i, \vr_i, \vpsi_i)\}_{i=1}^{N}}$ or used zero-shot with prompting.
While intuitive, this does not allow the model to explicitly reason about and verbalize a rationale of the reframing.
This is both less interpretable and might lead to overly-simplistic reframings.

Our proposed method \textsc{SocraticReframe} aims to overcome this by making the model rationalize the thought process behind a specific reframing by means of a sequence of Socratic question-answer pairs.
This means that we introduce an additional variable $\vs \in \mathcal{V}^\ast$ that is just a string of this question-answer sequence.
The model is then tasked to first generate $\vs$ and only then is allowed to generate $\vr$.
This is outlined in~\Cref{fig:full_framework}.
Hence, the model now becomes
\begin{equation}
\begin{split}
&p_\text{\vparam}(\vs\circ \vr \mid \vt, \vpsi) = \\ &\qquad \prod_{n=1}^{|\vs \circ \vr|}p_\text{\vparam}\big((s \circ r)_n \mid \vt, \vpsi, (\vs \circ \vr)_{<n}\big), 
\end{split}
\end{equation}
where $\vs\circ \vr$ means that $\vs$ is prepended to $\vr$.
For models that we finetune, this means that each training example is also augmented with a Socratic rationale $\vs$ such that $\smash{\data_{\text{train}} = \{(\vt_i, \vr_i, \vpsi_i, \vs_i)\}_{i=1}^{N}}$.
During inference, $\vs$ is not known and the model generates it before generating the positive thought.
This forces the model to explicitly reason about why a specific positive thought is generated to reframe a corresponding negative thought which we expect to improve both performance and interpretability.
In the following section, we discuss how such Socratic rationales can reliably be synthetically generated for existing datasets by leveraging LLMs.

\subsection{Generating Socratic Rationales}
\begin{figure}[t!]
    \centering
    \includegraphics[width=\columnwidth]{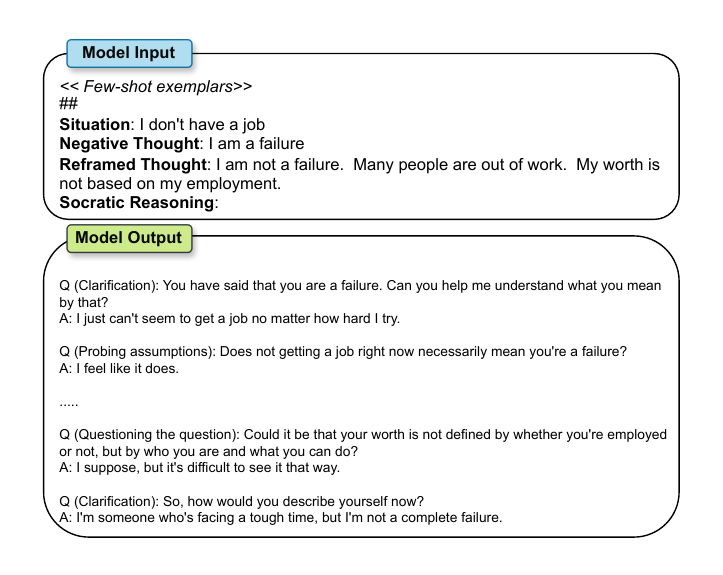}
    \caption{A sample Socratic rationale for an instance from COGREF \cite{sharma-etal-2023-cognitive} generated with GPT-4 in a few-shot setting. We use clinical vignettes from psychology literature as few-shot exemplars. 
    }
    \label{fig:socratic_example}
\end{figure}
Our goal is to improve cognitive reframing with Socratic rationales.
Such rationales can be time-consuming to write.
Therefore, we use synthetically generated data which has recently been successfully used in text classification \cite{li-etal-2023-synthetic}, dialogue generation \cite{bao-etal-2023-synthetic} and question-answering \cite{riabi-etal-2021-synthetic}, among others.

In particular, we few-shot prompt GPT-4~\citep{achiam2023gpt} to generate Socratic question-answer pairs based on negative thought $\vt$, positive thought $\vr$, metadata $\vpsi$, and in-context exemplars of Socratic rationales from clinical literature \cite{padesky1993socratic} to ensure high quality and groundedness.
The generated Socratic rationale $\vs$ then gives explicit reasoning steps for the particular reframing and can be used to augment the training data $\smash{\data_{\text{train}}}$.
This can be seen as a form of data augmentation and knowledge distillation into the model that is trained for cognitive reframing.
As we will show in~\Cref{human_evals}, the Socratic rationales are rated favourably by human annotators.
An example rationale is shown in \Cref{fig:socratic_example}. Details and examples are in~\Cref{sec:exp-details} and~\Cref{sec:app-examples}.
In addition to generating rationales, we prompt GPT-4 to classify the specific type of each generated Socratic question according to the six types described in \Cref{tab:socratic_types}, namely ``Clarification'', ``Probing Assumptions'', ``Probing reasons or evidence'', ``Probing implications'', ``Probing alternative viewpoints'', and ``Question''.
These can be used for further insights into the reframing process.

\section{Empirical Analysis} \label{experimental}
We show the effectiveness of our method on multiple datasets for positive text rewriting that are outlined in~\Cref{subsec:datasets}.
More details on the exact experimental set-up, such as the models used, are found in~\Cref{subsec:setup} and~\Cref{sec:exp-details} and the used metrics are introduced in~\Cref{subsec:metrics}.

\subsection{Datasets}
\label{subsec:datasets}
Table \ref{tab:datasets} describes the datasets used in this work which we further outline in the following.
\begin{table}[t!]
\centering
\small
\begin{tabular}{lp{1.5cm}p{1.5cm}p{1.5cm}}
\hline
 & {POSREF} & {PATREF} & {COGREF} \\
\hline
Train & 6,679 & 5,249 & 400 \\ 
Test & 835 & 18,635 & 200 \\ 
$\vpsi$ & - & Persona & Situation \\ 
\hline
\end{tabular}
\caption{Statistics of the used datasets.}
\label{tab:datasets}
\end{table}

We use three recently-released open-source datasets.
First, Positive Psychology Frames (POSREF) \citep{ziems-etal-2022-inducing} which contains tweets where a hashtag has indicated stress. 
Each tweet is mapped to a corresponding reframed text which is grounded in a set of reframing strategies, for example, Growth Mindset, Optimism, Self-affirmation. POSREF does not contain any metadata.
Then, we use Pattern Reframe (PATREF) \citep{maddela-etal-2023-training}, a crowdsourced dataset of negative thoughts that are conditioned on personas. These personas describe the person experiencing the respective negative thought. 
Here, $\vpsi$ contains the persona-specific information associated with each example. For instance, a persona from the dataset includes ``My mother was a teacher. My favorite food is a salad. I enjoy nature. I teach a yoga class. I am single.''
Finally, we use Cognitive Reframe (COGREF) \citep{sharma-etal-2023-cognitive} which is an expert-annotated dataset of situations, thoughts and reframes.
Mental health practitioners were prompted with a situation, which can be used as metadata, to generate a negative and a reframed thought. For example, one situation from the dataset is ``I participated in a hackathon and I lost''.
\begin{table*}[t!]
\centering
\large
\resizebox{\textwidth}{!}{

\begin{tabular}{c | c | c c c | c c c | c c c | c c c | c c c }
    \toprule 
     &  & \multicolumn{6}{c}{Content Preservation ($\uparrow$)} & \multicolumn{9}{c}{Transfer Strength ($\uparrow$)} \\ \cmidrule{3-8} \cmidrule{9-17}
     &  &  \multicolumn{3}{c}{BLEU} & \multicolumn{3}{c}{BLEURT} & \multicolumn{3}{c}{$\Delta$Pos} & \multicolumn{3}{c}{Acc. (\%)} & \multicolumn{3}{c}{$\Delta$Emp} \\ \cmidrule{3-17} 
    {Dataset} &  {Model} & {LLaMa} & {Mistral} & {ChatGPT} & {LLaMa} & {Mistral} & {ChatGPT}& {LLaMa} & {Mistral} & {ChatGPT}& {LLaMa} & {Mistral} & {ChatGPT}& {LLaMa} & {Mistral} & {ChatGPT} \\
    \midrule
    & FS & 12.6 & 13.1 & 15.1 & 0.53 & 0.55 & 0.57 & 0.54 & 0.49 & 0.54 & 89.15 & 90.10 & 92.00 & 0.58 & 0.66 & 0.70 \\ 
    & CoT  & 12.1 & 13.3 & 15.6 & 0.51 & 0.55 & 0.59 & 0.55 & 0.48 & 0.57 & 90.02 & 91.16 & 92.30 & 0.61 & 0.69 & 0.73 \\ 
     POSREF & FT & 14.3 & 15.1 & - & 0.56 & 0.58 & - & 0.55 & 0.51 & - & 91.46 & 91.10 & - & 0.63 & 0.72 & -\\
    & SoC$_{CoT}$  & 13.9 & 13.6 & 14.5 & 0.49 & 0.51 & 0.55 & 0.51 & 0.49 & 0.51 & 89.82 & 90.42 & 92.12 & 0.55 & 0.59 & 0.71\\ 
    \rowcolor[gray]{0.95}
    & SoC & 15.7 & \textbf{15.9} & - & 0.58 & \textbf{0.61} & - & \textbf{0.58} & 0.52 & - & \textbf{92.45} & 92.09 & - & 0.69 & \textbf{0.76} & -\\
    \midrule
    & FS & 71.9 & 73.0 & 73.8 & 0.64 & 0.65 & 0.63 & 0.49 & 0.78 & 0.81 & 97.02 & 96.51 & 97.61 & 0.81 & 0.80 & 0.87 \\ 
    & CoT  & 72.0 & 73.5 & 73.9 & 0.64 & 0.65 & 0.65 & 0.48 & 0.77 & 0.83 & 97.23 & 96.01 & 97.60 & 0.83 & 0.81 & 0.89 \\ 
     PATREF & FT & 72.3 & 74.1 & - & 0.66 & 0.67 & - & 0.51 & 0.80 & - & 97.67 & 96.79 & - & 0.89 & 0.85 & -\\
    & SoC$_{CoT}$  & 70.2 & 73.5 & 71.2 & 0.59 & 0.61 & 0.62 & 0.48 & 0.62 & 0.78 & 95.02 & 95.64 & 96.10 & 0.80 & 0.82 & 0.86\\ 
    \rowcolor[gray]{0.95}
    & SoC & \textbf{75.2} & 75.1 & - & \textbf{0.69} & 0.68 & - & 0.52 & \textbf{0.82} & - & \textbf{97.90} & 97.00 & - & \textbf{0.90} & 0.88 & -\\
    \midrule
    & FS & 27.1 & 27.9 & 30.0 & 0.59 & 0.61 & 0.63 & 0.67 & 0.68 & 0.69 & 90.12 & 90.94 & 91.90 & 0.79 & 0.81 & 0.88 \\ 
    & CoT  & 27.0 & 27.5 & 31.6 & 0.59 & 0.60 & 0.63 & 0.67 & 0.69 & 0.71 & 90.0 & 91.67 & 92.55 & 0.82 & 0.84 & 0.90 \\ 
     COGREF & FT & 29.0 & 30.0 & - & 0.61 & 0.62 & - & 0.69 & 0.70 & - & 91.48 & 91.88 & - & 0.89 & 0.91 & -\\
    & SoC$_{CoT}$ & 26.4 & 27.1 & 30.6 & 0.55 & 0.59 & 0.60 & 0.65 & 0.68 & 0.70 & 90.44 & 91.54 & 92.0 & 0.80 & 0.85 & 0.89\\ 
    \rowcolor[gray]{0.95}
    & SoC & 30.1 & \textbf{31.7 }& - & 0.62 & \textbf{0.63} & - & 0.70 & \textbf{0.72} & - & 92.30 & \textbf{92.56 }& - & 0.91 & \textbf{0.94} & -\\
    \bottomrule
\end{tabular}
}
\caption{Automatic evaluation results. FS=Few shot. CoT = Chain of Thought. FT=Finetune with LoRA. SoC=finetune with Socratic rationale. Items in \textbf{bold} represent the best performance. 
Note that we do not finetune ChatGPT.
We observe that using Socratic rationales for fine-tuning models significantly improves the text rewriting performance across all datasets. Mean values over three runs are reported. }
\label{Acc:all_models}
\end{table*}

\subsection{Experimental Set-up}
\label{subsec:setup}
We use different LLMs that we either prompt or finetune.
Namely, we prompt ChatGPT, LLaMa-2 7B \cite{touvron2023llama} and Mistral 7B \cite{jiang2023mistral} and
finetune LLaMa 7B and Mistral 7B for cognitive reframing.
All models are finetuned using LoRA \cite{hu2021lora} due to computational constraints.
For generating Socratic rationales, we use GPT-4.
We use the transformers \cite{wolf-etal-2020-transformers} library for all our experiments.

We test the models on different setups. \textit{Few Shot (FS):} Few shot exemplars of negative and positive thought are given to the model which are handcrafted from literature on cognitive reframing. \textit{Chain of Thought (CoT):} The model is asked to think step-by-step and rationalize the reframed thought. \footnote{We observed similar performance for zero-shot and few-shot CoT and therefore only report zero-shot CoT. We believe the similar performance in both setups stems from the nature of the reframing task where there is extra contextual information in the form of $\vpsi$ and the LLMs are able to generalize well even without few-shot exemplars.} \textit{Finetune (FT):} The model is finetuned using LoRA to reframe a negative thought without a Socratic rationale.
\textit{Socratic CoT (SoC$_{CoT}$):} The typical CoT prompt is modified by replacing ``think step-by-step'' with an instruction to generate Socratic rationale before generating the reframe.
\textit{Finetune with Socratic Rationales (SoC):}  The model is finetuned using LoRA to reframe a negative thought by first rationalising it using Socratic questioning and then generating the positive reframing. The prompt template used for finetuning is shown in Figure \ref{soc-prompt}.

\subsection{Evaluating Cognitive Reframers}
\label{subsec:metrics}
We aim to reframe a negative into a positive thought while preserving the original meaning and semantics. Hence, following \citet{hu-etal-2022-text,qu-etal-2023-conditioning}, we evaluate them along the criteria of Transfer Strength, which describes how well the thought was transferred to a positive one, and Content Preservation, which evaluates how well the original meaning of the thought was preserved.
We describe how we measure both of them in the following.

Transfer strength defines how well negative sentiment is turned  positive.
To measure it, we first use a finetuned RoBERTa model to evaluate the sentiment scores of the original and reframed thoughts.
Then, we use the pairwise difference of sentiment scores between the original and reframed thoughts and report the average, denoted by ({$\Delta$Pos}). We report the number of samples with an increase in positivity with respect to the original thought as accuracy. A decrease in the score from the original thought is considered a failed case ({Acc}). 
\citet{sharma-etal-2023-cognitive} show that people tend to prefer more empathetic reframes over overly positive reframes. Hence, we also report {$\Delta$Emp} as the difference between empathy scores of original and reframed thoughts. The scores are computed using a pretrained RoBERTa empathy classifier.
Our aim is to reframe a thought but also preserve the original meaning and not drastically change it. We rely on the following automatic metrics which have been widely used in the text reframing literature: {BLEU} \cite{papineni2002bleu} and {BLEURT} \cite{sellam-etal-2020-learning}. Following all three datasets we consider \cite{ziems-etal-2022-inducing,maddela-etal-2023-training,sharma-etal-2023-cognitive}, we report the BLEU scores for each pair of original and reframed thoughts. 
Since it has been shown that BLEU scores do not always correlate well with human judgements on semantic similarity, we also use a BERT-based BLEU variation, BLEURT-20 which is trained on synthetic samples to get accurate semantic similarity scores. We report the average over a dataset. 
Each score is a value between 0 and 1, ranging from no to complete semantic similarity.

\subsection{Evaluating Socratic rationales}
For Socratic rationales to be useful for a model, they should intuitively contain new salient information that is not contained in the negative thought.
We use the recently-proposed information-theoretic metric \textsc{REV} \cite{chen-etal-2023-rev} to measure this.
\textsc{REV} uses conditional $\mathcal{V}$-information to compute the usable information that can be extracted from a variable ($\vs$) by a model to predict another variable ($\vr$), conditioned on a third variable ($\vt, \vpsi$).  More formally, we compute
\begin{equation}
\begin{split}
    \textsc{REV}(\vt,\vr,\vpsi,\vs) &= -\log \smash{p_{\text{\vparam}^{\prime}}}(\vr\mid \vt,\vpsi) \\
    &\qquad + \log p_{\text{\vparam}}(\vr\mid \vt,\vpsi,\vs),
\end{split}
\end{equation}
where $\vparam^\prime$ and $\vparam$ are the parameters of two models trained to minimize cross-entropy, respectively.
The \textsc{REV} metric can then be computed for an entire corpus $\data$ by averaging the pointwise (per-example) scores.
While \citet{chen-etal-2023-rev} use fine-tuned models on the specific datasets for computing the score, \citet{lu-etal-2023-measuring} show that it can even be calculated directly from pretrained LLMs with the same effectiveness as using fine-tuned models \cite{lu-etal-2023-measuring}.

\label{subsec:socratic-rationales-results}
\begin{figure}[!t]
    \centering
    \includegraphics[width=\columnwidth]{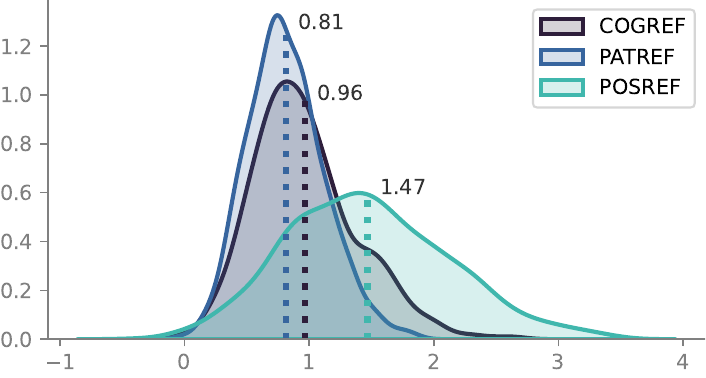}
    \caption{The Socratic rationales generated for all three datasets have positive $\textsc{REV}$ scores, meaning that they indeed provide useful information for text rewriting. Mean of the example-wise $\textsc{REV}$ values are plotted for all three datasets.}
    \label{fig:pvi-dist}
\end{figure}

\section{Results}
\label{results}
In this section, we detail our results obtained with different strategies.
First, our main results in~\Cref{subsec:main-results} show that the Socratic rationales improve text rewriting. Then, we show that the rationales are indeed informative for the model in~\Cref{subsec:socratic-rationales-results}.

\subsection{Main Results}
\label{subsec:main-results}
\Cref{Acc:all_models} shows that using Socratic rationales consistently outperforms both vanilla finetuning and prompting strategies on all datasets.
We show an example in Table \ref{tab:example_output} that highlights how the positive rewritings appear much more thoughtful, because the Socratic rationale forces the model to explicitly reason about the rewriting process.
We empirically check whether the generated reframes are close in length to the ground truth reframes for all datasets. We observe that the average length of generated reframes with our best performing model on all datasets is only 5 tokens longer than the ground truth, on average.
It is noteworthy that Socratic reasoning improves both the content preservation as well as transfer strength capabilities of smaller models beyond the performance of ChatGPT, despite the model being multiple times larger than LLaMa 7B and Mistral 7B. 
Across all datasets, we observe that Mistral performs better in content preservation while LLaMA performs better in the sentiment of the reframed thought. This could be due to the different pretraining datasets used for training the models. 
We also observe that our method consistently generates more empathetic reframes which is generally preferred by people \cite{sharma-etal-2023-cognitive}. We note that both positivity and empathy are lower for POSREF than other datasets, possibly because POSREF contains tweets which are known to be less positive in nature \cite{sokolova2017studying}.
We observe that using Socratic reasoning improves BLEU scores by almost 2 points on average when compared to finetuning without Socratic rationales. Distilling Socratic reasoning (from GPT-4) into smaller models helps to preserve the original meaning and improve the positive sentiment transfer. This is desirable for users and practitioners alike, because more positive sentiment will likely benefit users more and content preservation ensures that the reframing stays relevant for them.
\begin{table}[!t]
    \centering
    \resizebox{\linewidth}{!}{\begin{tabular}{p{2cm}p{6cm}}
    \hline
        {Method} & {Reframed Thought} \\ \hline
        Mistral-FS & I will improve. \\ 
        Mistral-CoT & I will improve and learn to be better. \\ 
        Mistral-FT & I will learn from the feedback and grow. \\ 
        Mistral-SoC& It is normal to feel disappointed. I can use this experience to learn and grow. \\ \hline
    \end{tabular}}
    \caption{Sample reframed thoughts generated by Mistral variants for the original thought: ``I submitted a paper to ACL and it got rejected. I will never succeed as a researcher.'' The additional socratic rationale leads to a more detailed and well-founded reframing.}
    \label{tab:example_output}
\end{table}

\subsection{Socratic Rationales are Informative}

Similar to previous work for evaluating free-text rationales \cite{chen-etal-2023-rev}, we use the pretrained GPT-Neo 2.7B \cite{gao2020pile} for computing \textsc{REV}. We report the average \textsc{REV} metric over each sample for all datasets.
A \textsc{REV} value > 0 suggests Socratic rationales support reframing by providing additional information, while < 0 indicates otherwise. In Figure \ref{fig:pvi-dist}, all datasets have \textsc{REV} > 0, highlighting Socratic rationale's utility in enhancing reframing tasks.
Table \ref{tab:generated_samples} shows the generated rationales for each of the dataset we consider. In particular, POSREF has the highest informativeness for the reframed thoughts. 
We attribute this to two main factors: First, POSREF lacks additional context like situations or personas, causing token dispersion where more information can lead to low values of \textsc{REV}. Second, POSREF samples from Reddit align better with GPT-Neo's training data than the crowdsourced data in the other datasets.

\subsection{Socratic Rationales progressively lead to a better reframe}
\begin{figure}[!h]
    \centering
    \includegraphics[width=\columnwidth]{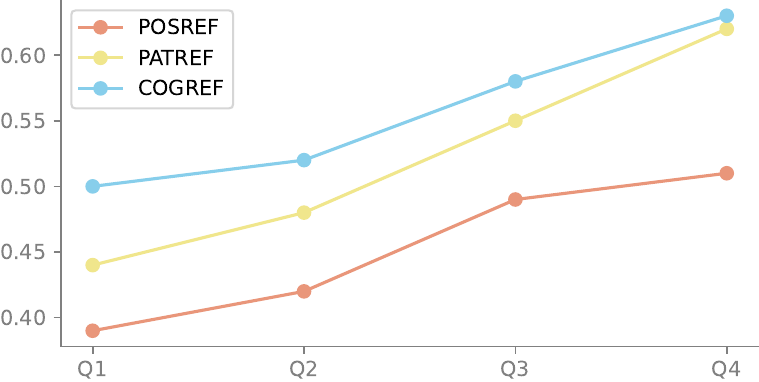}
    \caption{Sentiment scores for the generated Socratic question-answer pairs show that the synthetic rationales progressivley get better to lead to an improved reframe.}
    \label{fig:socratic_sentiment}
\end{figure}

We partition Socratic rationales into four quarters sequentially, each representing 25\% of the data, and calculate average sentiment scores for each quarter. \Cref{fig:socratic_sentiment} reveals a consistent trend across all datasets: intermediate answers lead to increasingly positive questions, suggesting that they aid in iterative reframing.

\section{Human Evaluation} \label{human_evals}

\subsection{Evaluating Reframed Thoughts}
Prior cognitive reframing studies \cite{maddela-etal-2023-training,sharma-etal-2023-cognitive} use Likert scales for assessing aspects like fluency and readability. Yet, our focus lies on helping users overcome unwanted thoughts.
\citet{li2024dissecting} show that pairwise human preferences are well-defined.
In a similar setup as \citet{sharma-etal-2023-cognitive}, we compare 100 randomly selected \textsc{SocraticReframe} generated reframes against ChatGPT or a baseline model without Socratic rationales. After consenting to participate, two computer science graduate students were recruited to conduct the comparisons. The win rate indicates a preference for Socratic reframes over the baseline by the raters. The participants were asked to select the reframed thought they find most relatable, helpful and memorable - a criteria for good reframes defined in \citet{sharma-etal-2023-cognitive}.

\begin{figure}
    \centering
    \includegraphics[width=\columnwidth]{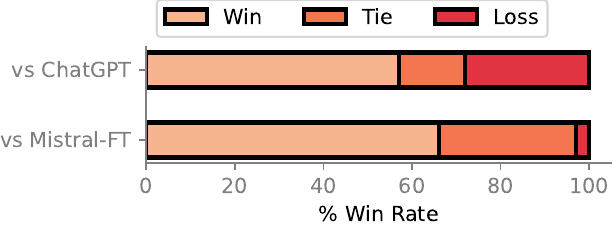}
    \caption{Human evaluation results for Mistral-SoC compared to Mistral-FT and ChatGPT. With a win rate of over 50\% human annotators prefer reframes generated by our method to ChatGPT and simple fine-tuning.}
    \label{fig:win-rate}
\end{figure}
As shown in Figure \ref{fig:win-rate}, Mistral-SoC outperforms Mistral-FT by a significant margin with a win rate of over 65\% and tie rate of around 30\%.
We fit a Bradley-Terry Model \cite{Bradley1952RankAO} on the preference data and obtain a strength of 1.3 for Mistral-SoC as compared to strengths -1.8 and 0.5 for Mistral-FT and ChatGPT, respectively. Intuitively, this means that Mistral-SoC will have a win rate of 95\% over Mistral-FT and 67\% over ChatGPT, confirming the effectiveness of our method.

\subsection{Evaluating Socratic Rationales}
To evaluate the synthetically generated rationales, we consider two key attributes: whether they represent the use of Socratic questioning in a clinically meaningful way and the general helpfulness of the questioning to overcome negative thoughts. 
Following~\citet{braun2015therapist}, we use the {Socratic Questioning Scale (SQS)} to evaluate the quality of generated rationales. 
SQS contains five Likert-scale questions evaluating the use of the Socratic method in a snippet. The final score is the sum of these ratings. Originally developed for therapy session transcripts, we include only three relevant questions for text evaluation: Open-endedness, Context, and Relevance.\footnote{{We note that in our evaluation criteria, ``relevant'' questions aim to tackle extrinsic hallucinations (generated text cannot be verified given the source) and ``contextual'' questions aim to tackle intrinsic hallucinations (generated text contradicts the source text), following the definitions from \citet{ji2023survey}.}} The full questions are reported in Appendix \ref{sec:sqs_protocol}. Additionally, we consider \emph{helpfulness} as a metric to assess whether or not the generated rationale is generally helpful in overcoming negative thoughts. Annotators rate the generated rationales on a 3-point Likert scale.

\begin{figure}
    \centering
    \includegraphics[width=\columnwidth]{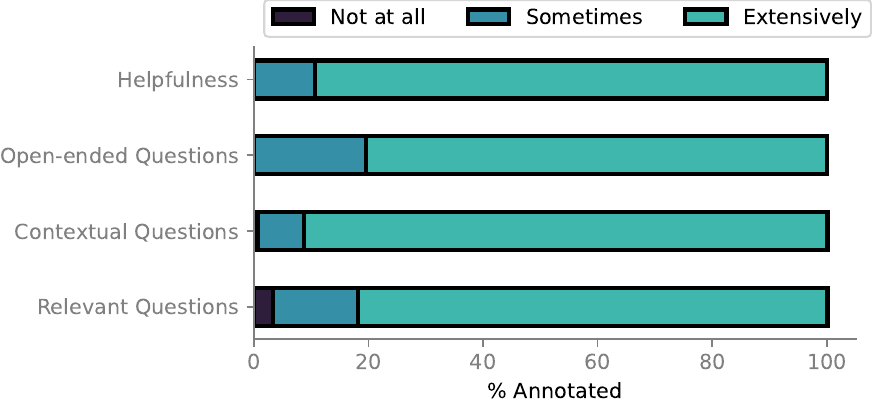}
    \caption{Human evaluation results for the Socratic rationales show that annotators judge the generated rationales as highly helpful and relevant for reframing negative thoughts.}
    \label{fig:sqs_ques}
\end{figure}

\paragraph{Expert Evaluation}
We recruit 14 mental health experts and clinical psychology graduate students for evaluating the quality of the Socratic rationales. We verify the quality of randomly sampled Socratic rationales with a clinical expert before conducting the study with psychology students. The annotation task was deemed exempt from an ethics review by an institutional review board.
\citet{amidei2019agreement} suggest the use of Cronbach's $\alpha$ for checking the evaluation reliability in natural language generation tasks. We observe acceptably high values for helpfulness (0.67), diversity (0.69), reflectiveness (0.75) and relevance (0.65) of the questions in the rationale. We observe that raters validate the synthetically generated Socratic rationales as helpful to reframe the original thought. More than 50\% of the raters also responded affirmatively on whether they are willing to use LLMs for cognitive reframing. More details are found in \Cref{validation-details}.

\paragraph{Human Evaluations} 
Three computer science graduate students annotated 100 randomly sampled Socratic rationales from the datasets. \Cref{fig:sqs_ques} shows favorable ratings across all criteria. We obtained an Intra Class Correlation (ICC) of 0.806 for helpfulness and 0.601 for the SQS scale, indicating that while the SQS is challenging to evaluate, helpfulness is generally agreed upon by non-experts. These findings align with similar agreement scores in psychology \cite{braun2015therapist}. Additionally, a Pearson correlation coefficient of 0.72 ($p<0.05$) between helpfulness and SQS ratings suggests that better Socratic questioning correlates with greater helpfulness in addressing negative thoughts.

\section{Conclusion \& Future Work} \label{conclusion}
In this work, we show that text rewriting models for cognitive reframing can be improved by using Socratic rationales to verbalize the reframing process.
By releasing our code and data we hope to enable future research, for example, on incorporating the different question types that are annotated, or using the data in therapist training.

\section*{Limitations}
While synthetic data generated by LLMs like GPT-4 offers a promising solution to data scarcity and privacy concerns, a drawback could be a potential lack of diversity and complexity. Our generated Socratic rationales may not fully capture all intricate nuances or patterns in authentic therapist dialogues while performing cognitive change with the Socratic method. 
A significant limitation when working with mental health and psychotherapy-based datasets is the use of heavily curated or crowdsourced data. This can lead to an imbalance in the demographics and language of our datasets. Since the datasets we use are sourced from social media or crowdsourcing platforms, the quality of the negative thoughts might also be limited.

In addition to the reframed thoughts, the datasets considered in this work also contain the cognitive distortions associated with each negative thought. Cognitive distortions are thought patterns that lead to negative feelings. We believe information like cognitive distortions and reframing strategies can be utilized to further improve the quality of the Socratic rationales and subsequently enhance the text reframing performance.

Finally, our focus in this work was to improve text reframing which is an effective short-term in-the-moment strategy. However, we emphasize that assessing long-term outcomes is imperative and is a future research direction.

\section*{Ethical Considerations}
While this work focused on generating Socratic rationales to improve positive text rewriting, open-ended LLMs still carry the risk of generating harmful outputs. Therefore, we advise careful consideration before they are applied in practice. Unsupervised use of the Socratic questioning data as is, without consulting trained professionals, could be harmful and requires careful design and implementation before being applied in real-world settings.


\bibliography{anthology1,anthology2,custom}

\newpage
\appendix

\section{Experimental details}
\label{sec:exp-details}
We few-shot prompt GPT-4 with the default parameters from the OpenAI Python package. \footnote{\url{https://github.com/openai/openai-python}} We use three few-shot exemplars.
Following similar works on synthetic data generation with LLMs \cite{macina-etal-2023-mathdial}, we use temperature sampling with $T=0.4$ and no top-k truncation for generating the reframes. 
For the fine-tuning experiments, following \citet{DBLP:journals/corr/abs-2305-14314}, we apply LoRA to all linear layers; training for 5 epochs with \textsc{AdamW}, a batch size of 8 and set the learning rate to 5e-4. All other hyperparameters take the default value in the configuration. We use NVIDIA A100 80GB for the training. 

\section{Prompt Design}
\subsection{Generating Socratic Rationales} \label{sec:gpt4-prompt}
\begin{tcolorbox}[colback=gray!5!white,colframe=gray!75!black, title=System Prompt for GPT-4]
   You are an expert in Cognitive Behavioral Therapy and Cognitive Restructuring, focusing on guided discovery.
The term cognitive restructuring refers to the process of challenging, and changing, irrational thoughts. Socratic questioning is one technique to encourage this process. Therapists use Socratic questioning verbally by asking probing questions about their clients' irrational thoughts. As clients improve their awareness of irrational thoughts, they can begin to consciously question their own thoughts.
The six types of Socratic Questions are: 1) Clarification, 2) Probing assumptions, 3) Probing reasons and evidence, 4) Questioning perspectives, 5) Probing implications and 6) Questioning the question. They can be used in any order and not all of them might be needed for a given client.
\end{tcolorbox}
\noindent\begin{minipage}{\columnwidth}
\captionof{figure}{This system prompt is used to prime GPT-4 with a personality grounded in the Socratic method.}\label{gpt-4-system-prompt}
\end{minipage}

An example of a clinical vignette taken from \citet{padesky1993socratic} used as a few-shot exemplar is:
\begin{lstlisting}
Negative Thought: "I'm a complete failure in every way."
Positive Thought: "I am not a failure."
Socratic Questioning:
Q (Clarification): You look defeated when you say that. Do you feel defeated?
A: Yes. I'm no good.
Q (Probing assumptions): You say you are no good. Is it true that you haven't done anything at all good?
A: Nothing of importance.
Q (Probing reasons and evidence): How about for your children this week -- did you care for them at all?
A: Of course, I helped my wife put them to bed and took them to soccer practice.
Q (Questioning perspectives): Do you think that was important to them?
A: I suppose so.
Q (Probing implications): And did you do anything to make your wife happy this week?
A: She liked the fact that I came home from work on time.
Q (Probing implications): Would a "complete failure" be able to respond to his wife's request in such a successful way?
A: I guess not.
Q (Probing implications): So is it really accurate to say you are a complete failure in every way?
A: I suppose not.
Q (Clarification): So how do you feel now?
A: I guess a little better.
\end{lstlisting}

The prompt template for GPT-4 is shown in Figure \ref{fig:socratic_example} and the system prompt is shown in~\Cref{gpt-4-system-prompt}.
\subsection{Finetuning}
\begin{tcolorbox}[colback=gray!5!white,colframe=gray!75!black, title= Prompt template used during finetuning,parbox=false]
Input:
Given a situation: \textsc{<<SITUATION>>} and the associated negative thought: \textsc{<<NEGATIVE THOUGHT>>}, generate the Socratic rationale for guided discovery and reframing the negative thought to a positive thought.
\tcblower
Output: \\
\textsc{<<SOCRATIC RATIONALE>>} \\
\textsc{<<POSITIVE THOUGHT>>}
\end{tcolorbox}
\noindent\begin{minipage}{\columnwidth}
\captionof{figure}{This template is used to finetune LLMs using the Socratic rationales.}\label{soc-prompt}
\end{minipage}
The prompt template used for finetuning is shown in~\Cref{soc-prompt}.

\section{Human Evaluation Protocol}
\subsection{Evaluating Socratic Rationales}\label{sec:sqs_protocol}
\paragraph{Socratic Questioning Scale (SQS)} The following questions were rated by annotators on a 3-point Likert scale ranging from 1 (not at all) to 3 (extensively) :
\begin{itemize}
    \item \textbf{How frequently were questions asked that help develop alternative perspectives?} - \textit{This question aims to assess the frequency of inquiries that encourage the exploration of diverse viewpoints or opinions. Annotators should focus on identifying questions that prompt respondents to consider different angles, challenge assumptions, or think beyond the conventional narrative.}
    \item \textbf{Was the question answering focused on the emotions and situation of the person?} - \textit{This question assesses whether the majority of answers provided focus on the emotional state and circumstances of the individual involved. Annotators should consider whether responses primarily address the feelings, experiences, or immediate context of the person in question.}
    \item \textbf{Were the questions open-ended and require thoughtful reflection?} - \textit{This question is designed to evaluate whether the questions posed to respondents are open-ended and demand deep contemplation rather than eliciting simple, direct answers. Annotators should look for questions that encourage respondents to think critically and provide nuanced, reflective responses.}
\end{itemize}

\paragraph{Helpfulness} Annotators rate the helpfulness of the question-answer pairs on a 3-point Likert scale ranging from 1 (not helpful at all) to 3 (very helpful). The following question was asked to the annotators:
\subparagraph{How helpful was the questioning in general?} \textit{This question seeks to gauge the overall effectiveness and utility of the questions posed. Annotators should consider the extent to which the questions contributed to a meaningful discussion, elicited insightful responses, or facilitated a deeper exploration of the topic. Assessments should encompass the clarity, relevance, and engagement level of the questions.}

We plan to release the generated rationales to complement the original datasets under the Apache license.\footnote{\url{https://www.apache.org/licenses/LICENSE-2.0}}.

\subsection{Validation with experts}
\label{validation-details}
We recruit psychology students for our study. The students were given credits in return of their annotation efforts. The average age of participants was 24.2 with 80\% participants being females. The participants were shown three diverse samples from across the datasets and asked to rate the helpfulness and SQS, in the same setting as described in \Cref{sec:sqs_protocol}. Additionally, participants were asked to answer the following question with either "Yes", "No" or "Not sure":
\subparagraph{Would you use an LLM to help you restructure negative thoughts?} \textit{This question assesses the willingness and openness of psychology students towards incorporating Large Language Models (LLMs) into their therapeutic practices. Responses provide insights into whether LLMs are perceived as viable tools for cognitive restructuring, indicating their potential acceptance within psychological interventions.}

We observe that only 7\% of the participants did not want to use LLMs for restructuring negative thoughts with 55\% responding affirmatively and the rest being unsure. We plan to conduct a more large-scale annotation study with more experts and non-experts in the future.
\subsection{Evaluating Reframes}
For a given context and negative thought, annotators are shown two sample reframes and asked the following question:
\begin{itemize}
    \item[] \textbf{Given the context and original negative thought, which reframed thought do you find more relatable, helpful and memorable (A vs B)?}
\end{itemize}

\section{Qualitative Examples} 
\label{sec:app-examples}
\Cref{tab:generated_samples} shows qualitative examples of generated socratic rationales for each of the three datasets used in our study.
\begin{table*}[!h]
    \centering
    \small
    \begin{tabular}{p{1.2cm}|p{5cm}|p{9cm}}
        \textbf{Dataset} & \textbf{Input} & \textbf{Generated Rationale} \\ 
        \toprule
        COGREF & \textcolor{gray}{Situation:} Having problems coming up with words to write essay. \par \textcolor{RedOrange}{Negative Thought:} I will get a bad grade. \par \textcolor{OliveGreen}{Reframed Thought:} I'm feeling anxious about this assignment. I can use skills to calm myself, try my best on the essay, ask for any appropriate help, and whatever grade I get I will know that I tried my hardest. & 
        \textcolor{gray}{Q:} You're worried about doing badly on this assignment. How have you done on past essays? \par
    \textcolor{gray}{A}: I've gotten good grades, sometimes. But this one feels different. \par
    \textcolor{gray}{Q:} Do you think getting a bad grade on this specific essay means you're generally bad at writing or at this subject overall? \par
    \textcolor{gray}{A:} Well, I don't think so, but I'm unsure about this topic. \par
    \textcolor{gray}{Q:} What evidence is leading you to believe that you will get a bad grade?  \par
    \textcolor{gray}{A:} I'm having trouble coming up with points to write about. \par
    \textcolor{gray}{Q}: If a friend was in your position, what advice would you give them?  \par
    \textcolor{gray}{A}: I would probably say to not panic and just give it their best shot. \par
    \textcolor{gray}{Q:} What would happen if you got a bad grade on this essay? How would that affect your overall academic achievement?  \par
    \textcolor{gray}{A:} One bad grade isn't going to ruin my overall performance I guess. \par
    \textcolor{gray}{Q:} Can using some support like talking to your teacher or a study group help you in getting through this? \par
    \textcolor{gray}{A:} That might be a good idea. \par
    \textcolor{gray}{Q:} So given these factors, do you still think it's a definite that you will get a bad grade? \par
    \textcolor{gray}{A:} No, I suppose it's not definite. \par
    \textcolor{gray}{Q:} So, what will you do about your essay now? \par
    \textcolor{gray}{A:} I think I'll start by focusing on writing a draft based on what I know and then ask for some help to see how I can improve it. \\ \hline
    PATREF & \textcolor{gray}{Persona:} my favorite band is rise against. my favorite team is the raptors. i usually go to bed late. i like sports. i am studying to be a dentist. \par \textcolor{RedOrange}{Negative Thought:} I am about to graduate and become a Dentist, but that means nothing.  Anyone can do that. \par \textcolor{OliveGreen}{Reframed Thought:} I am about to graduate and become a Dentist.  Although it is possible for anyone to become a dentist, it is very difficult, and I am proud of myself. & 
    \textcolor{gray}{Q:} It sounds like you're doing something remarkable, yet you downplay its value. Do you genuinely believe becoming a dentist means nothing? \par
\textcolor{gray}{A:} It doesn't seem special when everyone else is doing similar things.  \par
\textcolor{gray}{Q:} Do you think everyone could become a dentist? \par
\textcolor{gray}{A:} Maybe not everyone, but a lot of people could. \par
\textcolor{gray}{Q:} It required you to put in several years of hard study and practice, didn't it? \par
\textcolor{gray}{A:} Yes, I did work hard. \par
\textcolor{gray}{Q:} How many people do you think are willing or able to do what you did? \par
\textcolor{gray}{A:} I'm not sure, probably not that many. \par
\textcolor{gray}{Q:} So wouldn’t that mean that your achievement is indeed quite significant and not something anyone could do?\par
\textcolor{gray}{A:} I guess so.\par
\textcolor{gray}{Q:} Isn’t it possible then that your achievement of becoming a dentist is both valuable and hard-earned? \par
\textcolor{gray}{A:} Yes, it's starting to look that way.\par
\textcolor{gray}{Q:} How do you feel about your upcoming graduation now?\par
\textcolor{gray}{A:} I’m starting to recognize its importance and feel a bit proud.\par  \\ \hline
POSREF & \textcolor{RedOrange}{Negative Thought:} This physics homework is the death of me every week. It's so hard. And then I have another physics exam tomorrow night. \par \textcolor{OliveGreen}{Reframed Thought:} Physics is challenging, but is worth it. I have an exam tomorrow, which I hope to do well on. & 
\textcolor{gray}{Q:} It sounds quite challenging, is Physics always this difficult for you or only certain aspects? \par
\textcolor{gray}{A:} Certain parts, like quantum mechanics.\par
\textcolor{gray}{Q:} So, is it accurate to state that the entire subject is difficult, or just the quantum mechanics part? \par
\textcolor{gray}{A:} I suppose just the quantum mechanics part.\par
\textcolor{gray}{Q:} Can you give me an example of a physics problem or concept you've been able to understand quite well?\par
\textcolor{gray}{A:} Yeah, I've been doing well in classical mechanics.\par
\textcolor{gray}{Q:} Do you think that possessing skills in one area of physics could potentially aid in understanding other areas, including quantum mechanics? \par
\textcolor{gray}{A:} Maybe. But quantum mechanics seems very different.\par
\textcolor{gray}{Q:} If classical mechanics and quantum mechanics are two aspects of the same discipline, isn't there a possibility that understanding one could help with the other?\par
\textcolor{gray}{A:} It's a possibility, I guess.\par
\textcolor{gray}{Q:} Do you think saying that the physics homework is the death of you is a fair and accurate statement?\par
\textcolor{gray}{A:} Probably not, it's just really frustrating sometimes.\par
\textcolor{gray}{Q:} What if you thought about the physics homework as a challenge rather than a death sentence? Could that change how you approach it?\par
\textcolor{gray}{A:} Maybe. I could try. \par
\textcolor{gray}{Q:} Now, how do you feel about the physics exam tomorrow? \par
\textcolor{gray}{A:} I'm nervous, but maybe if I look at it as a challenge, it might feel less overwhelming.\par \\
\bottomrule
    \end{tabular}
    \caption{Socratic rationales generated with GPT-4 for different types of inputs from the three datasets we consider.}
    \label{tab:generated_samples}
\end{table*}

\end{document}